%% file: ofa2.tex
\documentclass[conference]{IEEEtran}
\usepackage[utf8]{inputenc}
\usepackage[T1]{fontenc}
\usepackage{graphicx}
\usepackage{longtable}
\usepackage{wrapfig}
\usepackage{rotating}
\usepackage[normalem]{ulem}
\usepackage{amsmath}
\usepackage{amssymb}
\usepackage{capt-of}
\usepackage{hyperref}
\IEEEoverridecommandlockouts
\usepackage{cite}
\usepackage{amsmath,amssymb,amsfonts}
\usepackage{algorithmic}
\usepackage{graphicx}
\usepackage{textcomp}
\usepackage{xcolor}
\usepackage{hanging}
\input{def.tex}
\input{author.tex}
\usepackage{xcolor,colortbl}
\usepackage[aboveskip=-0pt, belowskip=11pt]{subcaption}
\usepackage{atbegshi}
\AtBeginDocument{\AtBeginShipoutNext{\AtBeginShipoutDiscard}\addtocounter{page}{-1}}
\date{}
\title{OFA²: A Multi-Objective Perspective for the Once-for-All Neural Architecture Search}
\hypersetup{
 pdfauthor={Rafael Claro Ito},
 pdftitle={OFA²: A Multi-Objective Perspective for the Once-for-All Neural Architecture Search},
 pdfkeywords={},
 pdfsubject={},
 pdfcreator={Emacs 28.2 (Org mode 9.6)}, 
 pdflang={English}}
\makeatletter
\newcommand{\citeprocitem}[2]{\hyper@linkstart{cite}{citeproc_bib_item_#1}#2\hyper@linkend}
\makeatother

\begin{document}

\thanks{Part of the results presented in this work were obtained through the project “Hub of Artificial Intelligence in Health and Wellbeing – Viva Bem”, funded by Samsung Eletrônica da Amazônia Ltda., under the Information Technology Law no. 8,248/91}
\maketitle

\begin{abstract}
Once-for-All (OFA) is a Neural Architecture Search (NAS) framework designed to address the problem of searching efficient architectures for devices with different resources constraints by decoupling the training and the searching stages. The computationally expensive process of training the OFA neural network is done only once, and then it is possible to perform multiple searches for subnetworks extracted from this trained network according to each deployment scenario. In this work we aim to give one step further in the search for efficiency by explicitly conceiving the search stage as a multi-objective optimization problem. A Pareto frontier is then populated with efficient, and already trained, neural architectures exhibiting distinct trade-offs among the conflicting objectives. This could be achieved by using any multi-objective evolutionary algorithm during the search stage, such as NSGA-II and SMS-EMOA. In other words, the neural network is trained once, the searching for subnetworks considering different hardware constraints is also done one single time, and then the user can choose a suitable neural network according to each deployment scenario. The conjugation of OFA and an explicit algorithm for multi-objective optimization opens the possibility of a posteriori decision-making in NAS, after sampling efficient subnetworks which are a very good approximation of the Pareto frontier, given that those subnetworks are already trained and ready to use. The source code and the final search algorithm will be released at \url{https://github.com/ito-rafael/once-for-all-2}.
\end{abstract}
\begin{IEEEkeywords}
Neural Architecture Search,
Hardware-aware NAS,
AutoML,
Multi-objective Optimization
\end{IEEEkeywords}
\section{Introduction}
\label{sec:org2556e45}
Designing and training a deep neural network requires some user expertise and is often one of the most time-consuming steps of the machine learning pipeline. The field of AutoML aims to automate this process and started gaining some attention in recent years \citeprocitem{1}{[1]}. Specifically, the neural architecture search (NAS) subfield attempts to solve the problem of designing and searching a neural network given the demands of a learning task. This helps to alleviate the neural network designer from hand-crafting the architecture in a usually tedious trial and error procedure, while saving computational resources.

At the very beginning, the NAS algorithms were strictly focused on the automation process, leaving aside the computational resources concerns. In fact, the first NAS algorithms are supported by search strategies based on reinforcement learning \citeprocitem{2}{[2]} \citeprocitem{3}{[3]} or evolutionary algorithms \citeprocitem{4}{[4]} \citeprocitem{5}{[5]}, and the search cost took thousands of GPU days. Only later, when gradient-based search strategies were introduced, the computational burden was finally addressed \citeprocitem{6}{[6]}. The search cost was reduced to less than 10 GPU days, making it much more accessible for researchers to contribute to the field by improving the algorithms and proposing new ones.

However, with the recent progress in the IoT area and with the increasing amount of edge devices and edge computing, bringing machine learning into these devices may require the deployment of the application confined in different hardware settings, which is usually translated to hardware constraints such as latency, FLOPS and energy. Therefore, searching for an architecture for each specific device target leads to a computationally prohibited scenario, since training a single deep learning model can take weeks or even months to finish.
Besides not scaling well with the number of deployment scenarios, each additional training also presents an environmental burden. In fact, using NAS to search for a Transformer-like architecture \citeprocitem{7}{[7]} \citeprocitem{8}{[8]}, a popular architecture in the Natural Language Processing (NLP) area, is reported to cause as much as 5 lifetime cars in CO\(_2\) emission \citeprocitem{9}{[9]}.

\begin{figure}[!pb]
\centering
\includegraphics[width=1.00\linewidth]{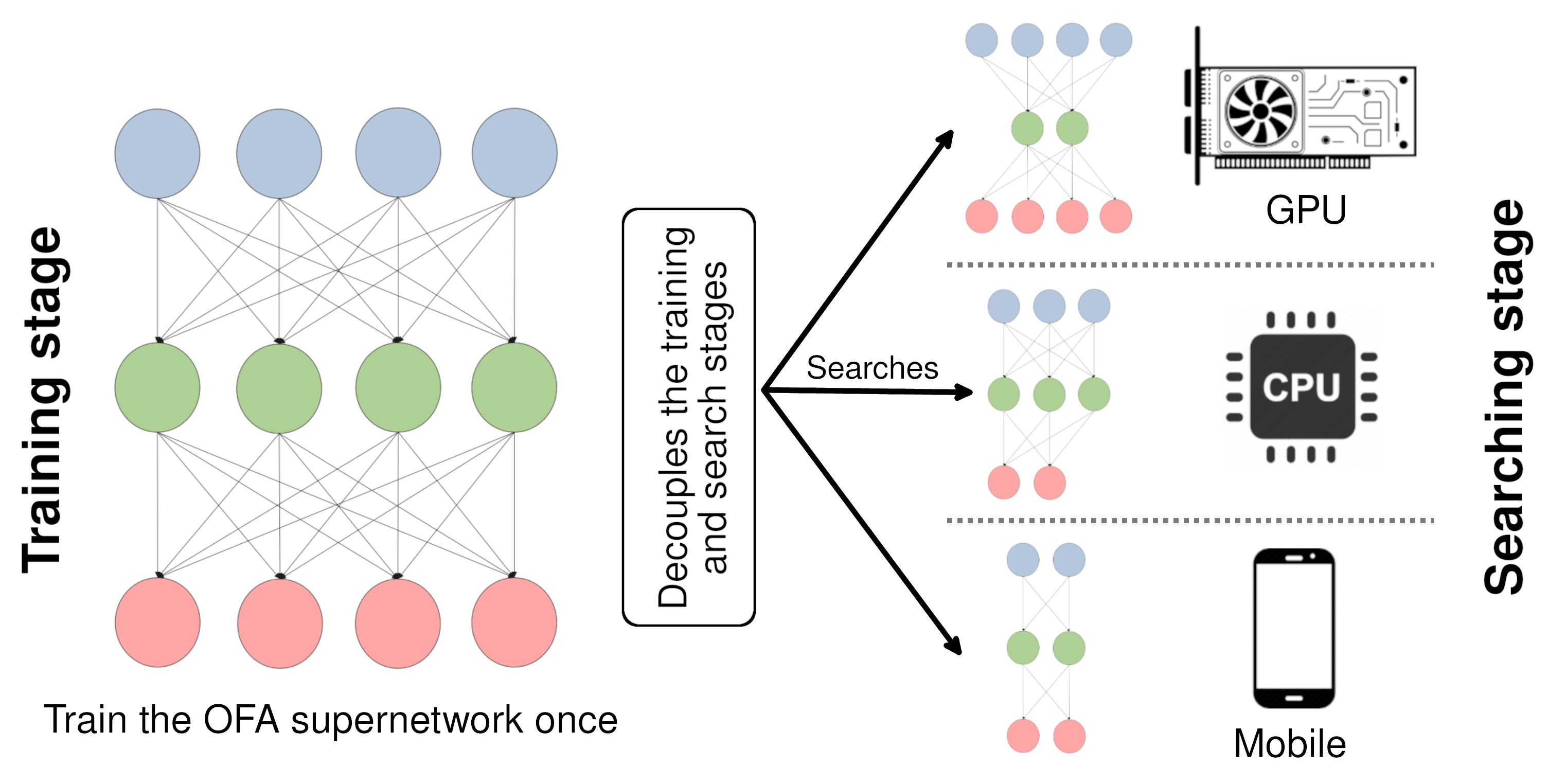}
\caption{\label{fig:ofa_overview}Once-for-All NAS framework overview}
\end{figure}
The Once-for-All (OFA) \citeprocitem{10}{[10]} is a NAS framework that attempts to address both the problem of having different deployment scenarios and the CO\(_2\) emission. This is done by decoupling the training stage from the search stage. In other words, this framework trains a single supernetwork once, and then performs multiple searches of subnetworks nested inside the already trained supernetwork for distinct deployment requirements. Since the training stage is the most costly step in the NAS process and it is done a single time, this cost can be amortized by the number of deployment scenarios. Figure \ref{fig:ofa_overview} illustrates the decoupling feature of the OFA framework.

\begin{figure}[!tpb]
\centering
\includegraphics[width=0.90\linewidth]{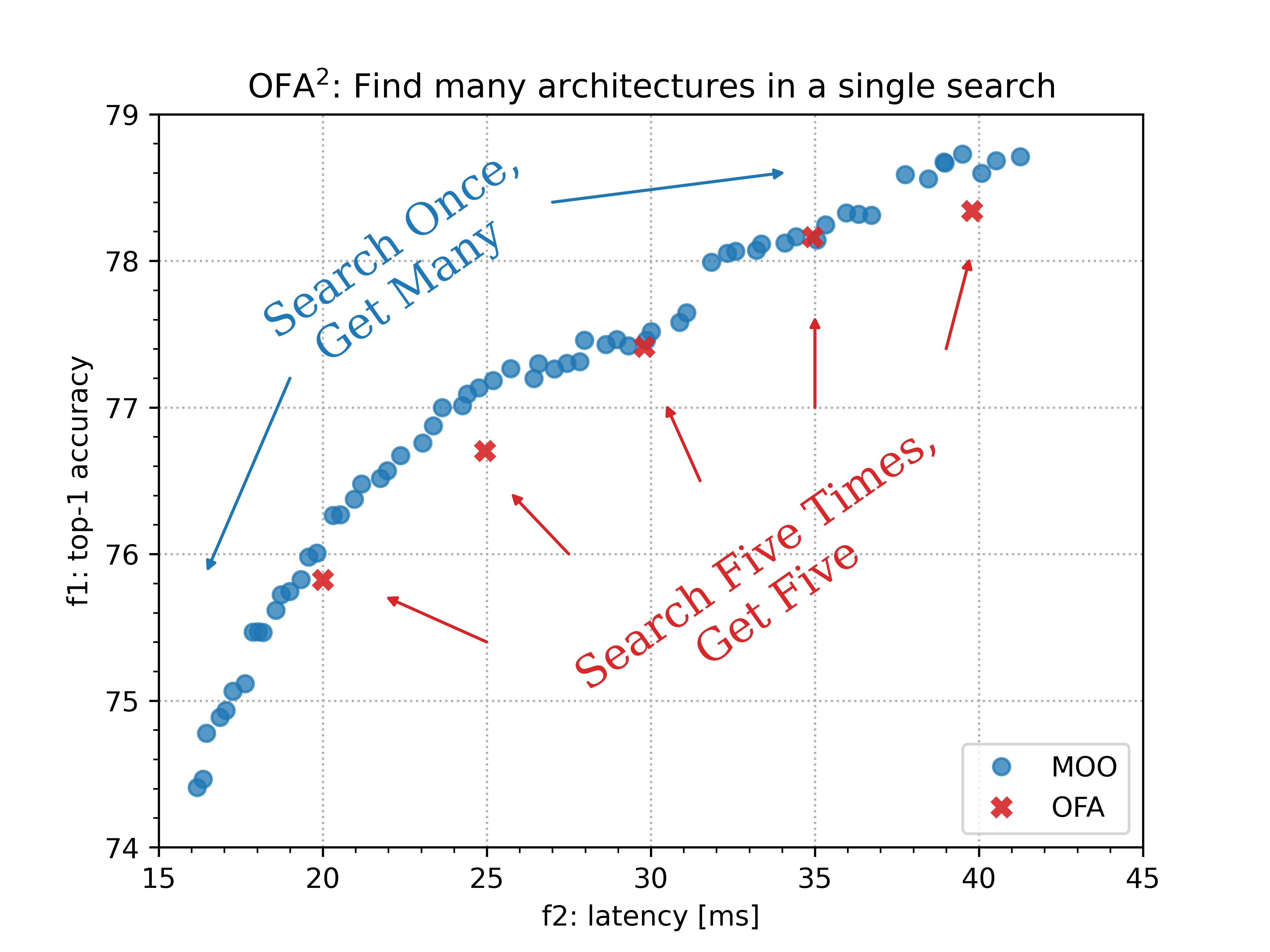}
\caption{\label{fig:search_once}Comparison between the search stage of the OFA and OFA\(^2\).}
\end{figure}

Although the search stage of the OFA framework is much cheaper than the training stage, it does not mean that the cost of this step is negligible. Moreover, the user must define a hardware constraint (latency or FLOPS) a priori to the search stage. This means that one search is required for each different deployment scenario, which is not very efficient. In this work we propose a multi-objective perspective to the search stage of the OFA framework, in such way that non-dominated neural networks for a representative amount of deployment scenarios are all produced by a single search procedure. This means that now the training stage of the OFA supernetwork is done once, but also the search stage of the OFA framework is done only a single time, and the output of this last stage now covers roughly all hardware trade-offs instead of a single constraint, as depicted in Figure \ref{fig:search_once}. Furthermore, we conduct experiments with those non-dominated neural networks as components of an ensemble, and compare performance against the single architectures and ensembles formed by architectures found by the original OFA framework and architectures randomly sampled from the OFA search space (without any search procedure).

We list the main contributions of this paper as follows:
\begin{itemize}
\item We extend the Once-for-All framework by proposing a multi-objective perspective for its search stage and solve this multi-objective optimization problem by finding the optimal architectures for different hardware constraints (e.g.: latency or FLOPS) all at once.
\item We compare the performance of architectures randomly sampled from the OFA search space, architectures found by the original OFA search algorithm and our multi-objective optimization method, and explicitly show the benefits in both accuracy and usage of computational resources compared to the others.
\item We conduct experiments with ensembles and we empirically show that choosing efficient architectures to form the ensemble beats the performance of ensembles composed of random architectures and also the ensembles formed by architectures returned by the original OFA search.
\end{itemize}
\section{Methods}
\label{sec:org63a962d}
Our work is an extension for the Once-for-All framework that aims to improve the search stage by selecting diverse architectures with better performance all at once. Since the training stage represents the step that demands most of the computational resources, we decided to take advantage of the fact that the OFA supernetwork was already trained and has its weights publicly available, and we focused on improving the search stage. More specifically, we formulated the search stage as a multi-objective optimization problem and solved it making the search more robust and leading to more efficient final architectures. Lately, we use the set of solutions obtained from the previous step to form ensembles and conduct a set of experiments comparing different scenarios.

The dataset used for both the training and searching stages is the ImageNet \citeprocitem{11}{[11]}, a standard dataset in the computer vision area that consists of 1,281,167 images in the training set and 50,000 images in the validation set, organized in 1,000 categories.

Next, we describe the training stage of the OFA framework just for completeness, and then we report the contributions of this work, including the changes in the search stage and the employment of efficient and diverse candidate solutions as components of state-of-the-art ensembles.
\subsection{Training stage}
\label{sec:org1c40f79}
The search spaces used in the NAS frameworks are typically divided into two categories: cell-based and one-shot.

While the cell-based approach consists in finding two cell structures (normal and reduction) in the form of a directed acyclic graph (DAG) and lately stacking these cells multiple times to form the final neural network architecture, the one-shot approach heavily relies on weight sharing, meaning that when one model inside this search space is trained, all other models that share these parameters will also have their weights updated (they are the same after all). The OFA framework presents a one-shot search space during the training stage.

The full OFA supernetwork is formed by 5 convolutional units placed in sequence. Each unit have 3 different parameters that can have their values changed: depth (number of layers) chosen from \{2, 3, 4\}, channel expansion ratio (number of channels) chosen from \{3, 4, 6\} and the convolution kernel size chosen from \{3, 5, 7\}. The fourth and last parameter that describes the search space is the resolution of the image that will be cropped from the original ImageNet and used as the input of the network. Its value is chosen from \{128, 132, \ldots{}, 224\}, totalizing 25 different input resolutions. Table \ref{tab:ofa_search_space} summarizes the OFA search space and the possible options to each hyperparameter. With the choices available, the amount of different subnetworks inside the OFA search space is approximately \(10^{19}\).

\begin{table}[!tpb]
\caption{\label{tab:ofa_search_space}OFA search space}
\centering
\begin{tabular}{c c c c}
\cellcolor{gray!30} PS & \cellcolor{gray!30} property & \cellcolor{gray!30} options & \cellcolor{gray!30} \# choices\\\empty
elastic resolution & input resolution & \{224, \ldots{}, 128\} & 25\\\empty
elastic kernel & kernel size & \{3, 5, 7\} & 3\\\empty
elastic depth & layers & \{2, 3, 4\} & 3\\\empty
elastic width & channels & \{3, 4, 6\} & 3\\\empty
\end{tabular}
\end{table}

\begin{figure}[!tpb]
\centering
\includegraphics[width=1.00\linewidth]{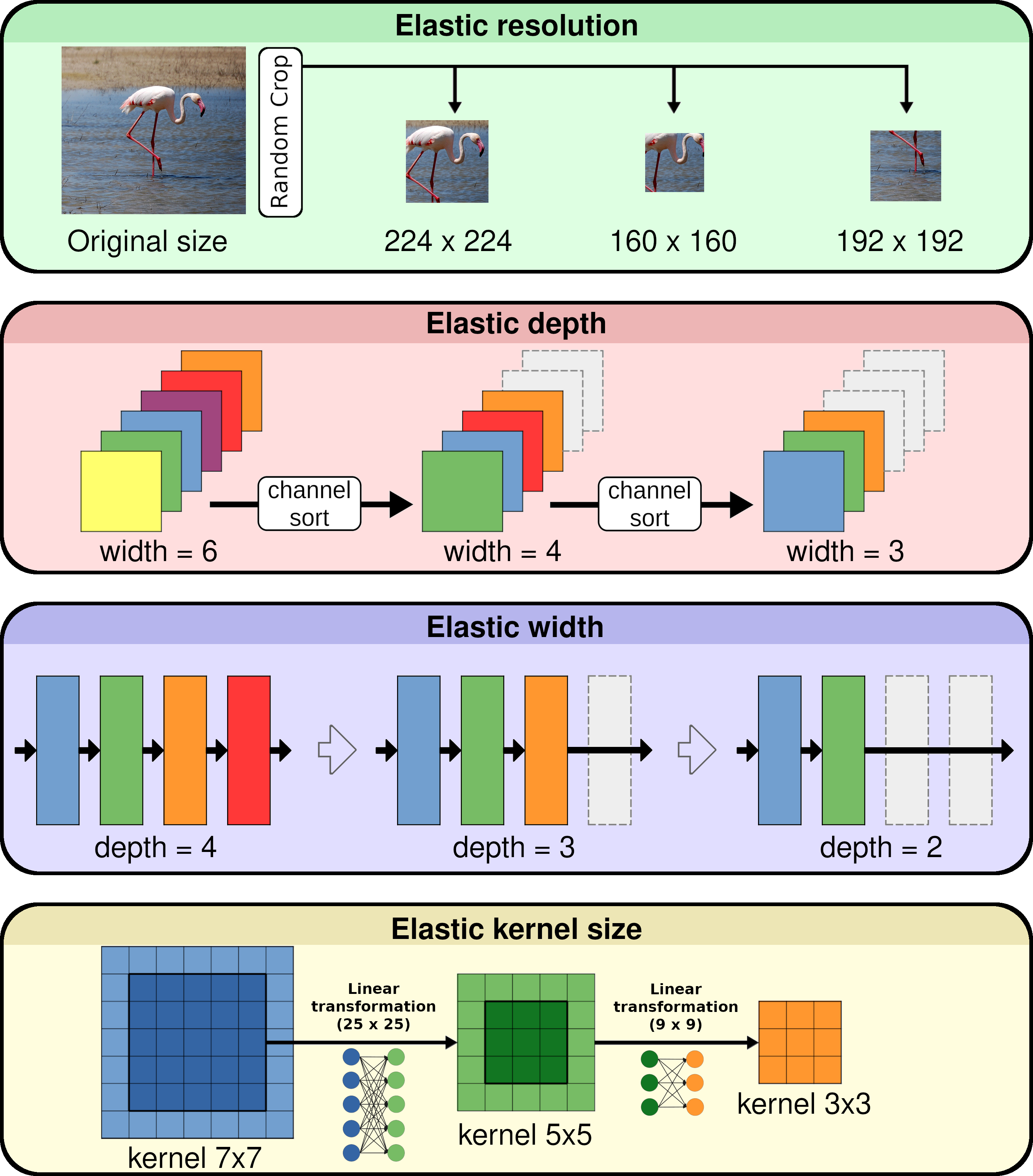}
\caption{\label{fig:ofa_ps}Progressive Shrinking (PS) algorithm.}
\end{figure}

The first step is to train the OFA network at its full capacity, that is, each of the 5 units having 4 layers, 6 channels with convolutions kernels of size 7x7. The resolution is randomly sampled during all the training. Then, after training the full network, the next step is to fine-tune the parameters training the subnetworks nested in the OFA supernetwork. This is achieved with the \emph{Progressive Shrinking} (PS) algorithm, introduced alongside the Once-for-All framework. The PS algorithm can be viewed as a general pruning technique, but instead of shrinking only a single dimension as it is usually done, it shrinks 4 dimensions gradually and in a sequence. It starts reducing the kernel size of the convolutional filters from 7x7 to 5x5, and then from 5x5 to 3x3, nesting the 3x3 kernel at the center of the 5x5 kernel, which is nested at the center of the 7x7 (elastic kernel size). Then skip connections are added to decrease the number of layers inside a unit (elastic depth), and finally it sorts the filters by its importance calculating the L1 norm of each channels' weights (elastic width). The main concepts are based on weights sharing, such that when the weights of a 3x3 convolutional kernel is being updated it also updates the 5x5 and 7x7 kernels, and also on the idea of sorting the subnetworks according to both accuracy and computational resources. Figure \ref{fig:ofa_ps} illustrates the progressive shrinking algorithm. The shrinking process is completely done in one hyperparameter at a time before moving to the next one (vertical axis of Figure \ref{fig:ofa_ps}) and always starts with the hyperparameters options that makes the neural network larger towards smaller architectures (horizontal axis of Figure \ref{fig:ofa_ps}). For example, with respect to the depth, the network is first trained with 4 layers, then with 3 and lately with 2. Please refer to the original paper for more detailed information about the training.

A final remark is related to the \emph{neural-network-twins}. Since evaluating the architecture accuracy on the validation set of the ImageNet dataset and measuring the latency of a model in a specific hardware can be expensive, the authors built an accuracy and a latency predictor as substitutes to speed up the search stage. The accuracy predictor is a three-layer feedforward neural network with 400 hidden nodes in each layer and the latency predictor is built as a lookup table \citeprocitem{12}{[12]} one for each different hardware being considered.
\subsection{Search stage}
\label{sec:org5d79fb4}
The original framework performs an evolutionary search on the trained OFA network to find an architecture with the meeting requirements based on an input provided by the user. This means that the resource constraints must be defined a priori to the search process and that the output of the framework is a single architecture. Here we propose the search process in a multi-objective perspective \citeprocitem{13}{[13]}, where the goal is to minimize two conflicting objective functions that represent the top-1 error and a hardware constraint (latency or FLOPS).

We tested two multi-objective optimization (MOO) algorithms in our experiments, both being evolutionary and population-based: the NSGA-II \citeprocitem{14}{[14]}, which uses the concepts of non-dominated sorting and crowding distance sorting to solve the problem, and the SMS-EMOA \citeprocitem{15}{[15]}, which uses the hypervolume metric to perform the evolution, besides the non-dominated sorting concept as well. The whole problem was modeled using a multi-objective optimization framework called pymoo \citeprocitem{16}{[16]}, which uses the NumPy library as its backend \citeprocitem{17}{[17]}. For the evaluation of the neural network architectures, a Titan X GPU with 11 GB of memory was used and the code was implemented using the PyTorch framework \citeprocitem{18}{[18]}.

\subsubsection{Genotype}
\label{sec:org870bc8b}
\begin{figure}[!tpb]
\centering
\includegraphics[width=1.00\linewidth]{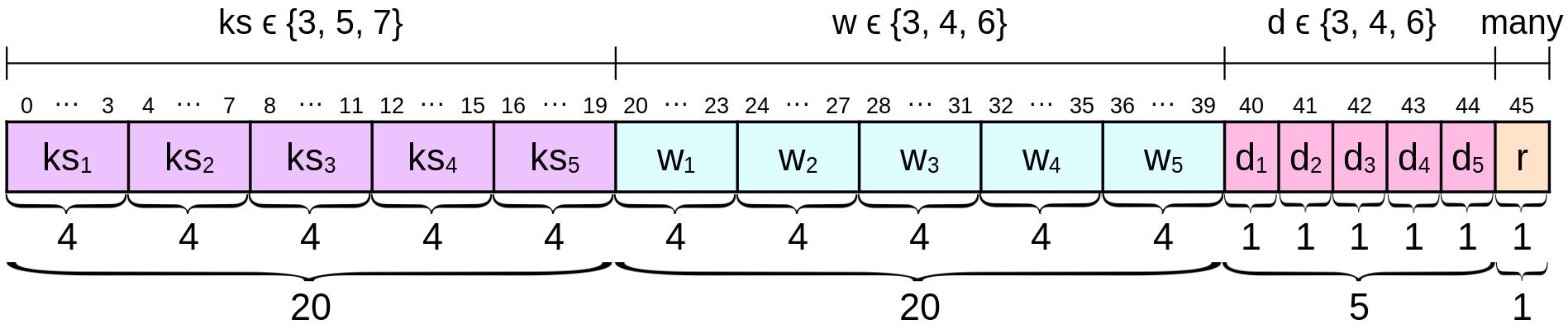}
\caption{\label{fig:moo_encoding}Genotype encoding used to represent an individual of the population.}
\end{figure}

The first step to solve the search stage as a MOO problem is to define the genotype encoding that will be used by the evolutionary multi-objective algorithm (EMOA) to represent an individual of the population, where each individual represents a full neural network and the population is just a set of neural networks considered at each iteration of the algorithm. For this, we simply flatten the hyperparameters used to represent an architecture from Table \ref{tab:ofa_search_space} in a one dimensional array. Since the final architecture is formed by 5 units in sequence and each unit may have up to 4 layers, we have a total of 20 entries to represent the convolutional kernel size (\texttt{ks}) and 20 entries to represent the width (\texttt{w}). We also have 5 entries related to the depth (\texttt{d}) of each unit showing the number of layers considered, and one last entry informing the resolution (\texttt{r}) of the image that will be cropped from the dataset and used as the input of the neural network. This gives us a total of 46 variables for the encoding of each individual that represents a full neural network. Figure \ref{fig:moo_encoding} illustrates this representation of an individual.

\begin{figure}

\begin{subfigure}{0.48\textwidth}
\begin{center}
\includegraphics[width=1.0\linewidth]{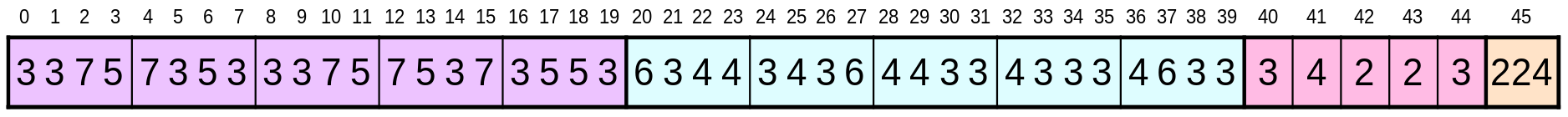}
\end{center}
\caption{\label{fig:moo_encoding_example_a}Numerical example of an individual.}
\end{subfigure}

\begin{subfigure}{0.48\textwidth}
\begin{center}
\includegraphics[width=1.0\linewidth]{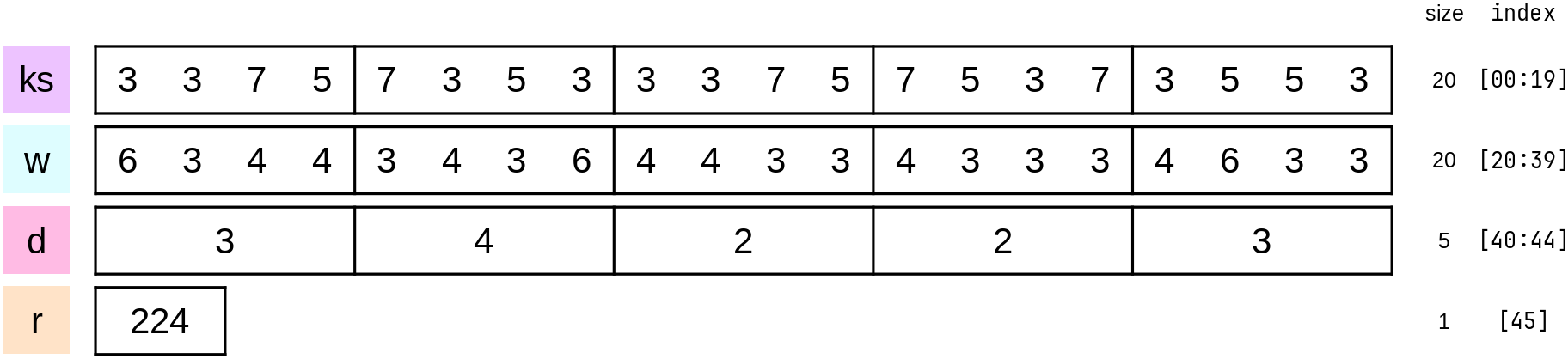}
\end{center}
\caption{\label{fig:moo_encoding_example_b}Split the encoding according to each hyperparameter.}
\end{subfigure}

\begin{subfigure}{0.48\textwidth}
\begin{center}
\includegraphics[width=1.0\linewidth]{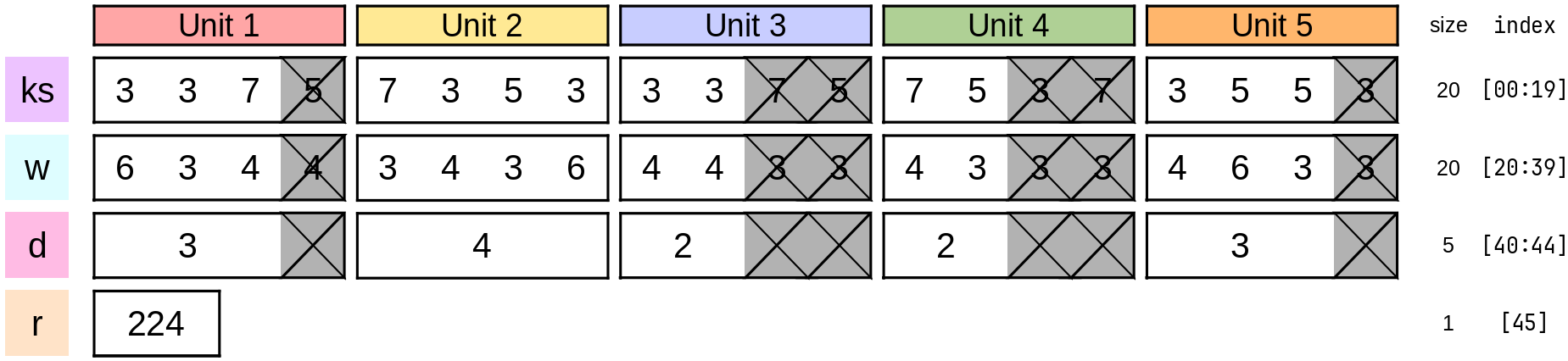}
\end{center}
\caption{\label{fig:moo_encoding_example_c}Division in units discarding data when depth \(\in\) \{2,3\}.}
\end{subfigure}

\begin{subfigure}{0.48\textwidth}
\begin{center}
\includegraphics[width=0.50\linewidth]{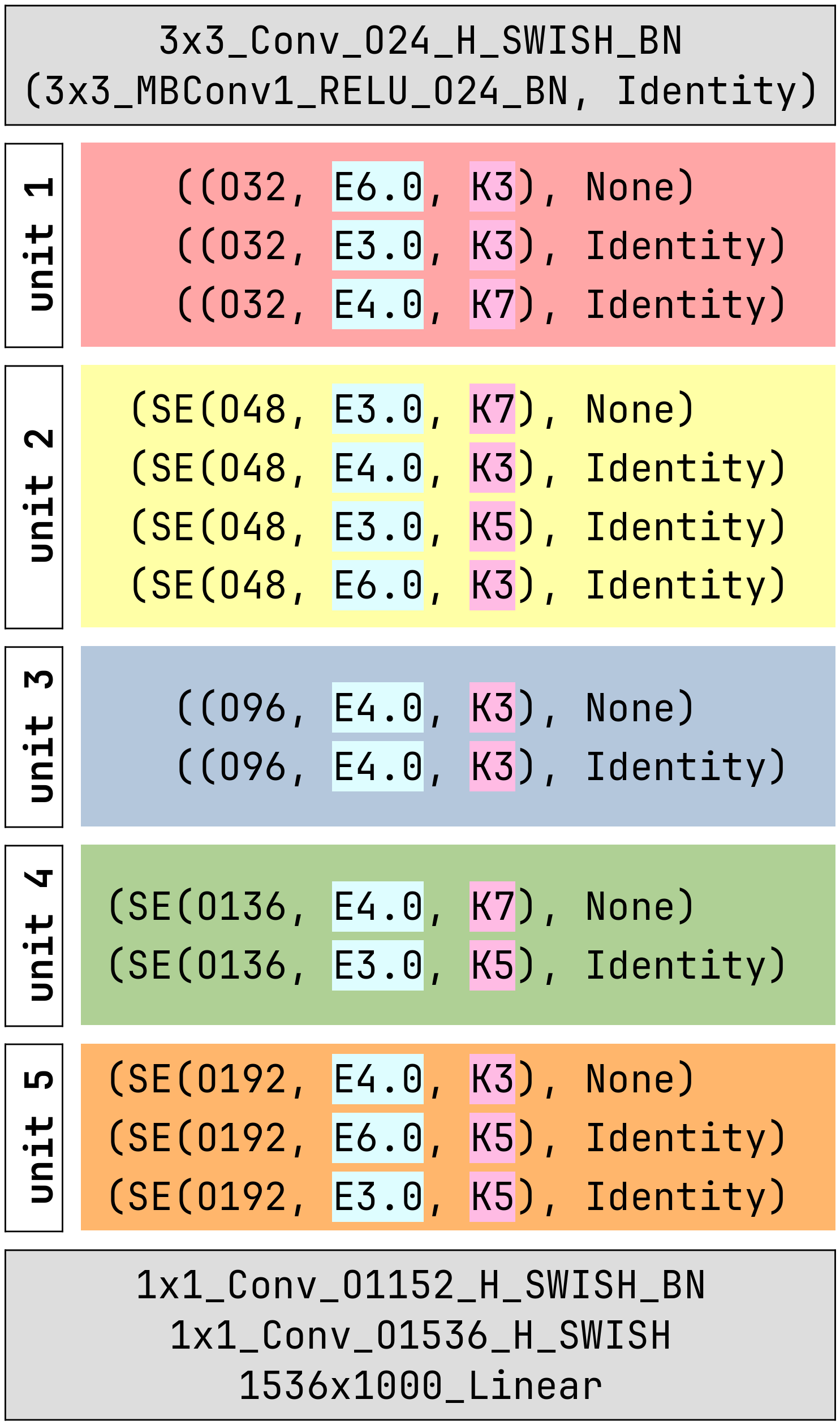}
\end{center}
\caption{\label{fig:moo_encoding_example_d}Final architecture representation.}
\end{subfigure}
\caption{\label{fig:moo_encoding_example}Encoding of an individual of the population.}
\end{figure}

Given the encoding of an individual, in order to build the full architecture associated with that individual, the first step is to separate each variable according to the hyperparameter it represents, following the scheme depicted in Figure \ref{fig:moo_encoding}. Figure \ref{fig:moo_encoding_example_a} illustrates a numerical example of an individual and Figure \ref{fig:moo_encoding_example_b} shows the respective division of its values according to the hyperparameters associated. After that, we can group these hyperparameters per unit. If the depth of the unit being considered is equal to 4, i.e. the unit has 4 layers, then all 4 values of the kernel size and all 4 values of the width are valid and will be used to build the model. If the depth of the unit is equal to 3 layers, then we simply discard the last value of both the kernel size and width. If the depth is equal to 2, then we discard the last 2 entries of both kernel size and width. Figure \ref{fig:moo_encoding_example_c} illustrates this procedure. Finally, with all hyperparameters values of the 5 units, we can build the full architecture, as shown in Figure \ref{fig:moo_encoding_example_d}.

\subsubsection{Objective Functions}
\label{sec:org583a9bf}
The objective functions we want to optimize is the top-1 accuracy (maximize) and another conflicting objective, represented by a resource constraint, such as latency of FLOPS (minimize). For this, we use the accuracy and latency predictors provided by the OFA framework. That is, given the encoding of an architecture we use the accuracy predictor (the same regardless the hardware) to estimate the model accuracy, and we use specific latency lookup table for each hardware to get the predicted latency. There are a few hardware options available to choose the latency lookup table, such as the Google Pixel 2 mobile phone, the NVIDIA GTX 1080 Ti GPU, CPU or even FLOPS. We chose the Samsung Galaxy Note 10 for our experiments. After obtaining the final population of architectures using a MOO algorithm with these predictors, we took each of the final models and performed an evaluation under the validation set of the ImageNet. During the evaluation, the resolution of the squared image was randomly chosen from \{160, 176, 192, 208, 224\}.
\subsubsection{Operators}
\label{sec:org2712b99}
Next, we define the four operators that will take place on the individuals during the iterations of evolutionary algorithms: sampling, mutation, crossover and selection. The first three operators are the same in both multi-objective optimization algorithms used to find the final architectures of the OFA search stage (NSGA-II and SMS-EMOA, discussed in the next section). The sampling operator is related to the initialization of the algorithm, that is, it defines the initial population of individuals at iteration zero. In our case we simply used the random sampling, meaning that values for the depth, width and kernel size were randomly chosen among their respective valid choices (\texttt{d} \(\in\) \{2, 3, 4\}, \texttt{w} \(\in\) \{3, 4, 6\}, \texttt{ks} \(\in\) \{3, 5, 7\}). The mutation operator is used to promote diversity among the solutions, which might prevent the algorithm to get stuck in a local minimum. In our experiments we defined that each gene of the chromosome has a probability of 10\% of replacing its value into one of the valid choices (including the same value), which is the same probability used by the evolutionary search proposed on the OFA framework. The crossover operator, also known as recombination, takes two solutions as parents and combine them to generate a child solution. Here we chose the uniform crossover, which means that the value of each gene of the child solution is randomly taken from one of the parents solution with equal probability. Finally, the selection operator defines a criterion for choosing the individuals of the current population that will be used to generate the offspring, that is, the next generation of individuals. It usually incorporates the fitness function (objective functions) somehow, and in our case this operator is implicitly defined according to which of the two MOO algorithms is used during the evolutionary search.
\subsubsection{MOO Algorithms}
\label{sec:org85d71a9}
The first MOO algorithm considered is the Non-dominated Sorting Genetic Algorithm II (NSGA-II) \citeprocitem{14}{[14]}. In this algorithm the selection operator chooses first the individuals based on their level of non-dominated front. When the sum of the individuals already selected with the amount of individuals of the current front being considered surpass the population size previously defined, then the individuals of this last front are selected based on their crowding distance (Manhattan distance in the objective space). The selection based on the rank of the front always favors the best solutions with respect to the objective functions, while the crowding distance selection aims to spread the solutions toward regions less explored. In our experiments, we defined the population size to be 100 and ran the algorithm for 1,000 generations.
The second MOO algorithm considered is the SMS-EMOA \citeprocitem{15}{[15]}. This algorithm guides the search aiming to maximize the hypervolume measure (or s-metric), which is commonly applied to compare the performance of different evolutionary multi-objective optimization algorithms (EMOA). This algorithm combines both the concepts of non-dominated sorting and the hypervolume measure as the selection operator. Similarly to the previous scenario, we used a population size of 100 and ran this algorithm for 1,000 generations.
It is important to note that others MOO algorithms could have been used in this stage.
\subsection{Ensemble}
\label{sec:org0eaa55e}
After the search stage we end up with a final population where each of the individuals represent an efficient architecture considering different trade-offs between the objective functions, i.e., accuracy and latency. We then realize a series of experiments grouping different individuals to form ensembles. There are three scenarios considered for the experiments, regarding the group of neural architectures in which the components of the ensemble will be taken from.
\subsubsection{Random components}
\label{sec:orge6a3c0e}
This group of architectures represents 100 neural networks sampled from the OFA search space without any search procedure. We randomly choose the value of each gene (depth, kernel size, expansion ratio) among its valid options. The procedure to get these architectures is the same of the one to define the initial population in the MOO algorithms.
\subsubsection{OFA components}
\label{sec:orga0a8d73}
The second group of architectures is the one obtained from the evolutionary algorithm of the original OFA framework. Since this search is guided by a specific hardware constraint, we need to perform a full search for each requested architecture, resulting in 9 runs of the search algorithm for the 9 architectures in this group. The restrictions used as input of the searches are the latencies from 15 ms up to 55 ms, increasing 5 ms at each step.
\subsubsection{Efficient components}
\label{sec:org571f7af}
The last set of architectures considered in the ensemble experiments are the architectures obtained from the MOO algorithm, which aims to optimize both accuracy and latency. Since the evolutionary algorithm is population-based, all architectures are found at once at the end of the search procedure. Each of them is optimal considering a trade-off between the conflicting objectives.

\begin{figure}[!pb]
\centering
\includegraphics[width=1.00\linewidth]{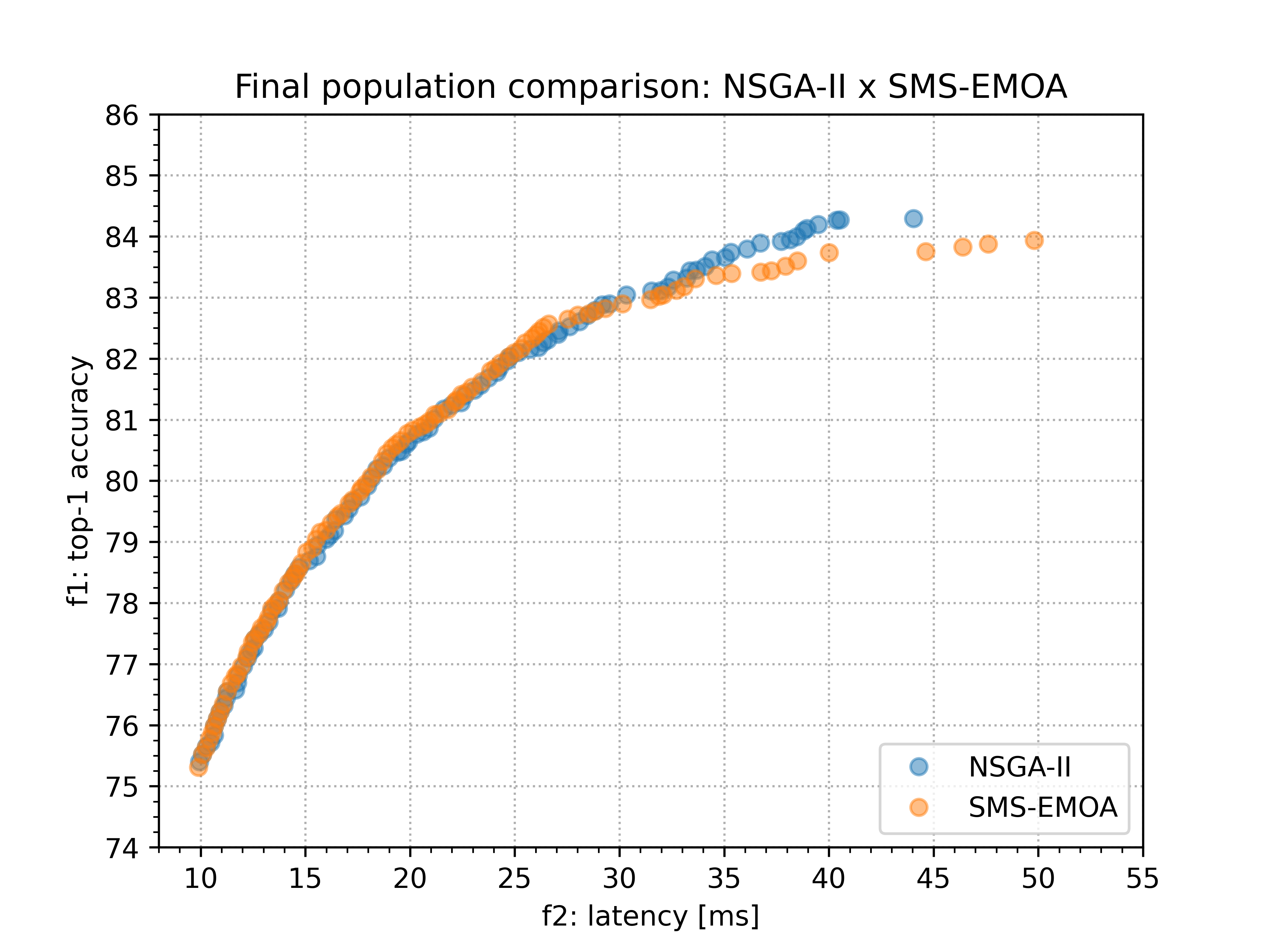}
\caption{\label{fig:nsga2_smsemoa_moo}Comparison between the NSGA-II and SMS-EMOA final population.}
\end{figure}
\subsubsection{Voting schemes}
\label{sec:orgc8e3625}
In order to evaluate the performance of the ensemble, we use two voting schemes. In the first one, called ``hard voting'', the output of the ensemble is decided according to the most-voted top-1 class among the participants of the ensemble. If there is a draw in votes between one or more classes, then we check the occurrences of these classes on the second most likely output of each model (top-2 output) and decided by the most frequent. If there is still a draw in votes, then we keep checking the top-3 up to the top-5 output. After that, if the draw still exists, we look at the architectures who voted on the classes with the same amount of votes on the top-1 output and choose the output of the biggest model to be the output of the ensemble (according to the premise the larger models have more flexibility and might be more reliable than smaller models). In this scheme, the output of each neural network has the same importance, regardless how certain the model is about its output.
The second voting scheme also take into account the probability assigned to each class on the output. For this, we take the last layer of each neural network and append a softmax layer straight after it. To decide the output of the ensemble, we sum the probabilities for each class among all participants of the ensemble, and take the output with highest accumulated value. This helps to alleviate one problem of the hard voting scheme, which is the fact that a vote from a model with a low confidence in its output has the same weight of a vote from a model with a high confidence. This second voting scheme, called ``soft voting'', provides a way to weight the vote of each architecture according to the confidence of the model in its output class, which can be beneficial in some cases.
\section{Results and Discussion}
\label{sec:org36b31b5}
\subsection{Searching for efficient architectures}
\label{sec:org9b1f730}
To solve the multi-objective optimization problem formulated for the search stage of the Once-for-All framework, we proposed using two MOO algorithms: NSGA-II and SMS-EMOA. Figure \ref{fig:nsga2_smsemoa_moo} shows the final architectures found by each of these algorithms after the 1,000 generations. We can see that the final population approximates the typical Pareto front for MOO problems with two conflicting objective functions. Moreover, it looks that while the SMS-EMOA finds slightly better architectures with lower latency, the NSGA-II finds better architectures from the middle up to the end of the latency axes. Appart from the region with higher latency, the predicted accuracies found by both algorithms are very similar though.

\begin{figure}[!pb]
\centering
\includegraphics[width=0.90\linewidth]{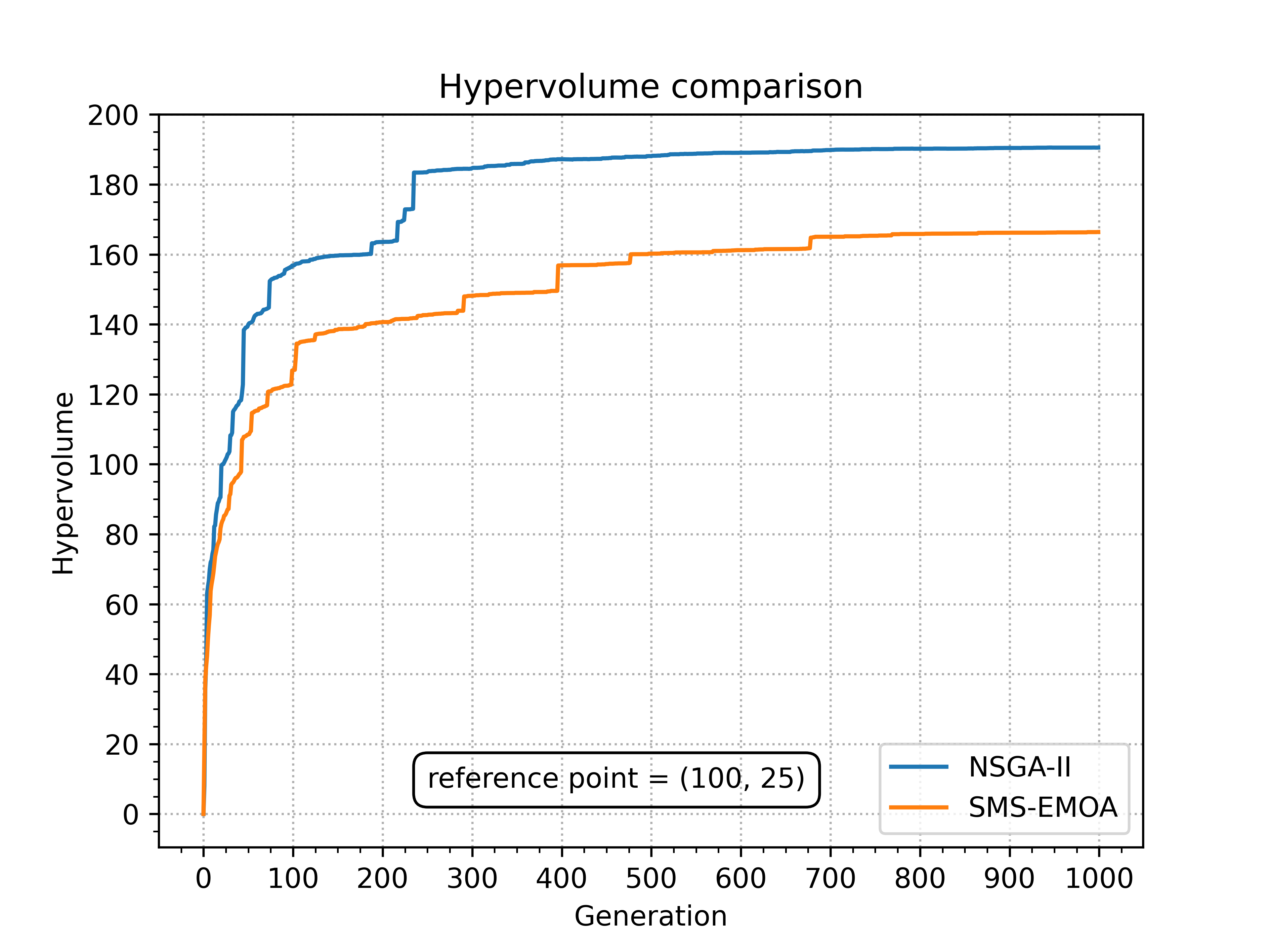}
\caption{\label{fig:nsga2_smsemoa_hypervolume}Comparison between the NSGA-II and SMS-EMOA hypervolume.}
\end{figure}

In Figure \ref{fig:nsga2_smsemoa_hypervolume} we plot the progression of the hypervolume measure (s-metric) of the NSGA-II and the SMS-EMOA across the 1,000 generations. This metric is commonly used to compare the performance of different evolutionary multi-objective optimization algorithms (EMOA) and requires a reference point on the objective space to be calculated, which for the Figure in question is the point (100, 25).

\begin{figure}[!tpb]
\centering
\includegraphics[width=1.00\linewidth]{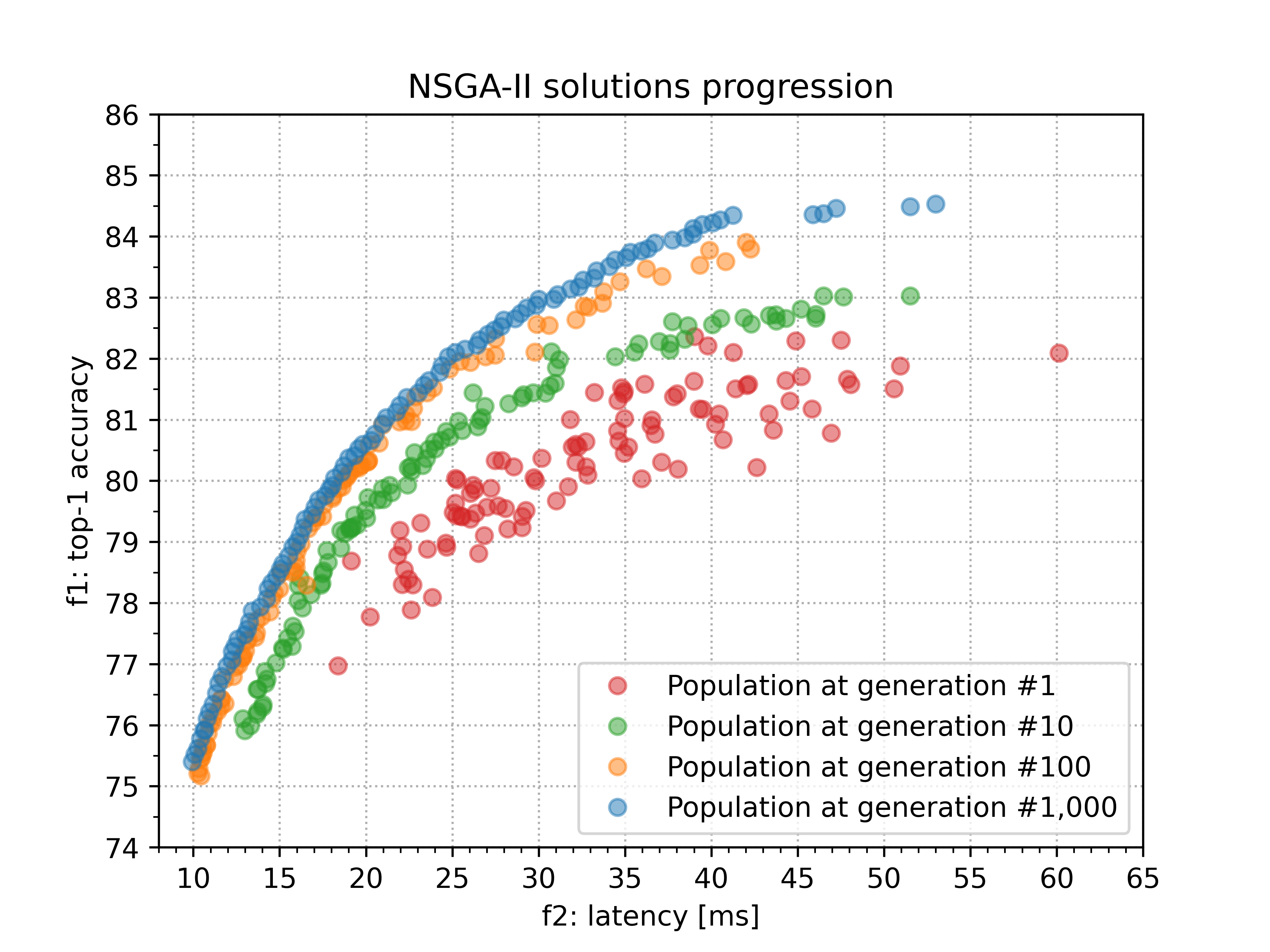}
\caption{\label{fig:nsga2_progression}Progression of the solutions for the NSGA-II MOO algorithm.}
\end{figure}

Once the NSGA-II presented a higher value for the hypervolume measure after the 1,000 generations of the evolutionary search, we decided to use the population found by this algorithm for the following experiments with the ensembles. Figure \ref{fig:nsga2_progression} shows the progression of the populations across the iterations of the algorithm. We can see that the individuals of the initial population (red) are spread across the objective space, which is comprehensible, since these individuals are randomly sampled from the OFA search space. Even though they are far from being efficient in terms of both accuracy and latency compared to the optimal solutions, over the generations they progressively approximate the Pareto front, as we can see for the individuals after 10 (green), 100 (orange) and 1,000 generations (blue). This indicates that even though the neural networks of the OFA search space have their weights already trained, a search is indeed needed to retrieve efficient architectures.

\begin{figure}[!tpb]
\centering
\includegraphics[width=1.00\linewidth]{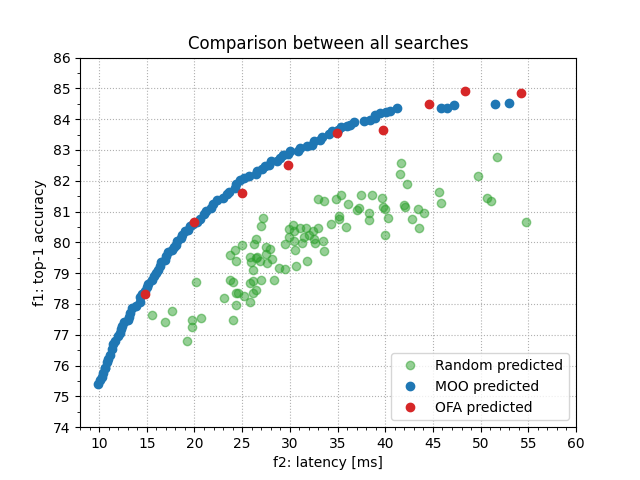}
\caption{\label{fig:comparison_pred_all}Comparison between the different search methods using the accuracy predictor.}
\end{figure}

Next, in Figure \ref{fig:comparison_pred_all} we show the comparison between the 100 individuals of the final population searched with the NSGA-II algorithm in a MOO approach, the 100 individuals randomly sampled from the OFA search space and the 9 individuals obtained with 9 runs of the original evolutionary algorithm of the OFA framework for 9 different latency constraint (15 ms to 55 ms, with a step of 5 ms). It is clear here the advantage of proposing the search process as a multi-objective optimization problem, since all solutions are found in a single search, given the user the power to make a choice a posteriori on which architecture to select.

The performance predictors are useful to speed up the search process. However, if the predictions are not accurate, the resulting architectures of the search process will not present the same results during inference. In order to check the reliability of the accuracy predictor, we took each of the architectures shown in Figure \ref{fig:comparison_pred_all} and measured the real accuracy under the validation set of the ImageNet. The results are shown in Figure \ref{fig:comparison_pred_all}. We can see that although the top-1 \% accuracies are not exactly the same, the shape of curve is, which means that the accuracy predictor works well disregarding a constant offset. Luckily, the domination concept used by the MOO algorithms are not dependent on the absolute values with respect to the objective functions, relying instead on a comparative approach, which means that as long the shape of the curves on the objective space is the same, we should not have any problem with these types of evolutionary algorithms.

\begin{figure}[!tpb]
\centering
\includegraphics[width=1.00\linewidth]{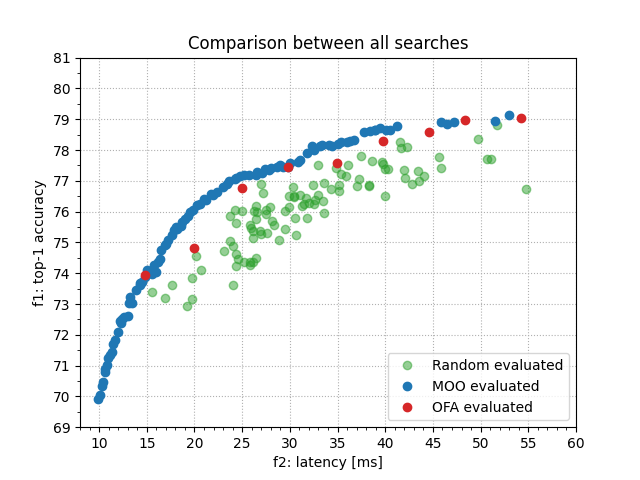}
\caption{\label{fig:comparison_eval_all}Comparison between the different search methods considering the real accuracy evaluated on the ImageNet validation set.}
\end{figure}

\subsection{Ensemble}
\label{sec:orgd5a6c47}
With the architectures represented in Figures \ref{fig:comparison_pred_all} and \ref{fig:comparison_eval_all}, we now perform a series of experiments grouping these neural networks to form ensembles and to compare the impact of each search method on the results of the committee machines. For each population we realize experiments considering both the hard and soft voting mechanisms. Furthermore, instead of taking only the accuracy of the ensemble into account, we also consider two approaches regarding the latency metric of the ensembles. In the first approach, the latency of the ensemble is defined to be the sum of all architectures participating in the ensemble. This strategy is based on the premise that we have a single hardware with limited amount of memory to implement the ensemble, and therefore we need to load each model one at a time to evaluate its performance. In the second approach, the latency of the ensemble is equal to the model's latency that has the highest value among those networks participating of the ensemble. This strategy is based on the premise that parallelization is viable, and therefore all models can be evaluated simultaneously. This, of course, requires a limited amount of models in the ensemble, due to memory scaling. As a consequence, we test ensembles with the number of components varying from 2 up to 8 (totalizing 7 different ensemble size) due to this limitation.
\subsubsection{Efficient components}
\label{sec:org89af546}
\begin{figure}[!tpb]
\centering
\includegraphics[width=1.00\linewidth]{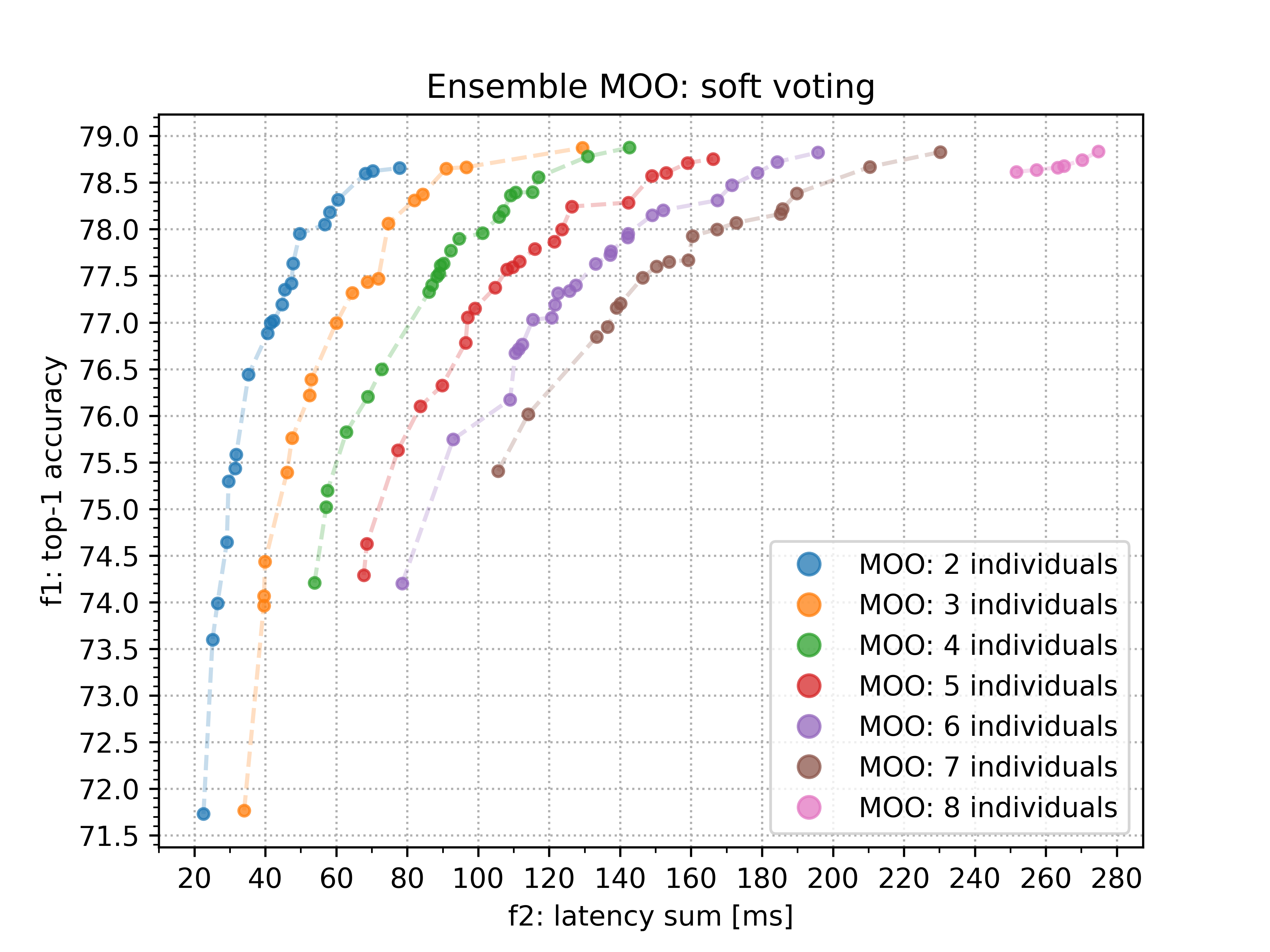}
\caption{\label{fig:ensemble_all}Non-dominated ensembles with different sizes composed by efficient architectures.}
\end{figure}

The first group of architectures considered in our experiments with ensemble are those of the last population obtained from the NSGA-II search described before, and we call it the efficient population. We then propose two different sampling techniques to form the ensembles. For both of them, the first step is to sort the neural networks of this population by latency, which means that the model with index 0 is the model with the lowest latency (most left model in Figure \ref{fig:comparison_eval_all}) and the model with index 99 is the one with the highest latency (most right model in Figure \ref{fig:comparison_eval_all}). In the first method, we sample 43 random different combinations of these architectures to make up the committee. We start with ensembles containing 2 architectures, then we sample more 43 random different combinations for ensembles with size 3, so on and so forth, up to ensembles composed by 8 neural networks. This method will be used as a comparison to the performance of the ensembles with components from the random architecture's population. The second approach uses a subset of the efficient population and aims to mimic the architectures found by the OFA search. For that, we take the already sorted by latency architectures of the efficient population and choose the architectures immediately before the latency constraint of 15 ms, then 20 ms, up to 55 ms. This approach will be put side by side of the ensemble formed with the architectures found with the OFA search. Figure \ref{fig:ensemble_all} shows all non-dominated architectures for the first method considering the sum of latencies and the soft voting scheme.

\subsubsection{Random components}
\label{sec:org00f9202}
\begin{figure}[!tpb]
\centering
\includegraphics[width=1.00\linewidth]{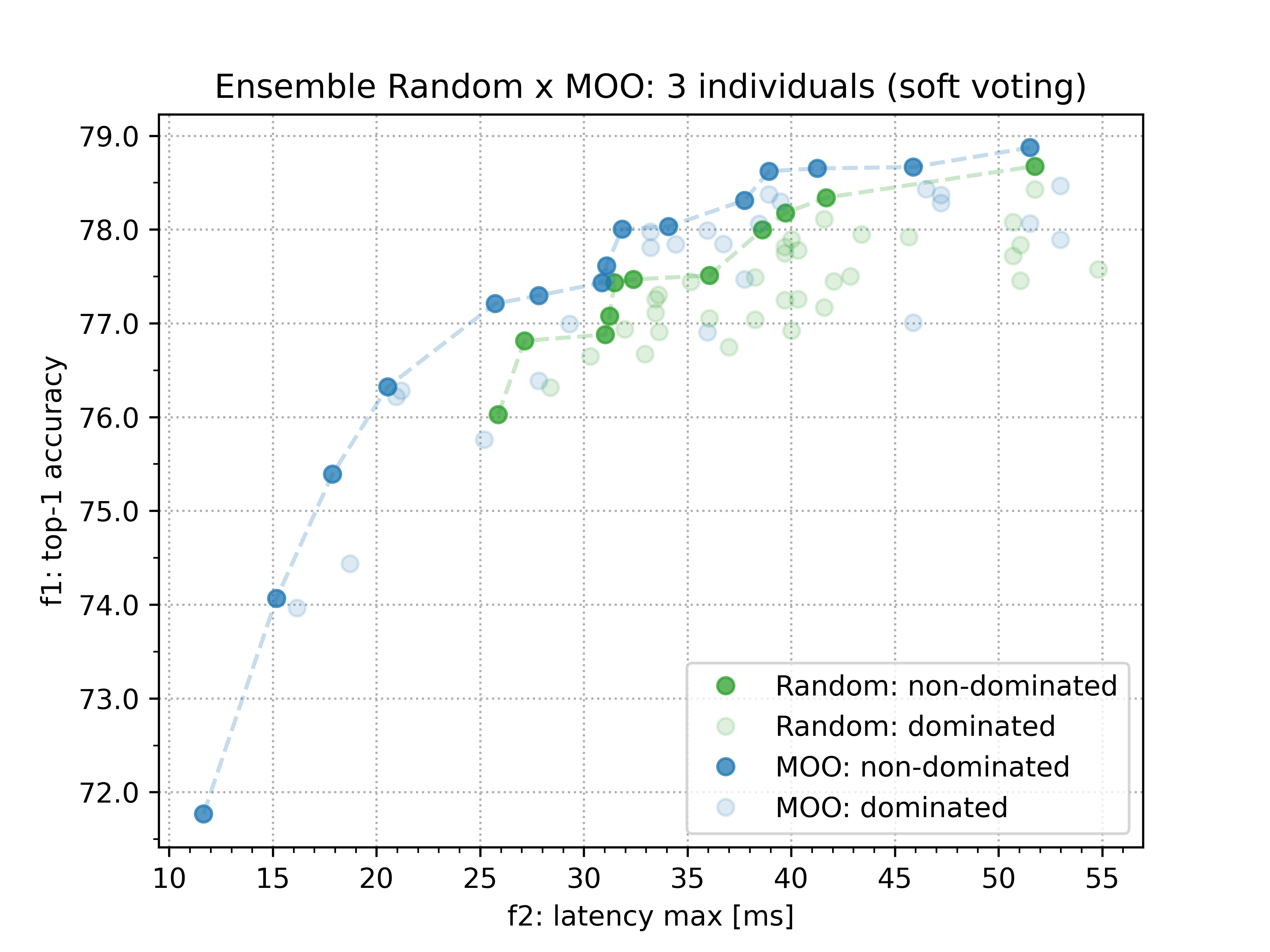}
\caption{\label{fig:ensemble_moo_random}Comparison between ensembles with efficient and random architectures (individuals = 3, latency = max, voting = soft).}
\end{figure}

The experiments with ensembles formed by the population of random architectures follow the first method of the ensembles with efficient architectures. That is, we take 43 different combinations of architectures for each of the ensembles with 2, 3, \ldots{}, 8 models in its compositions out of the 100 models available. All of this is done once for the hard voting scheme and once for the soft voting scheme, both with the sum and maximum of latencies in the ensemble. Figure \ref{fig:ensemble_moo_random} shows an example of comparison between the ensemble with efficient (blue) and random (green) architectures, using the soft voting scheme and considering the highest latency on the committee. We can see that the ensembles found using the efficient architectures consistently outperforms the ones found with random architectures. This same statement is valid for roughly all other experiments related to these two populations.

\subsubsection{OFA components}
\label{sec:org54d711a}
\begin{figure}[!tpb]
\centering
\includegraphics[width=1.00\linewidth]{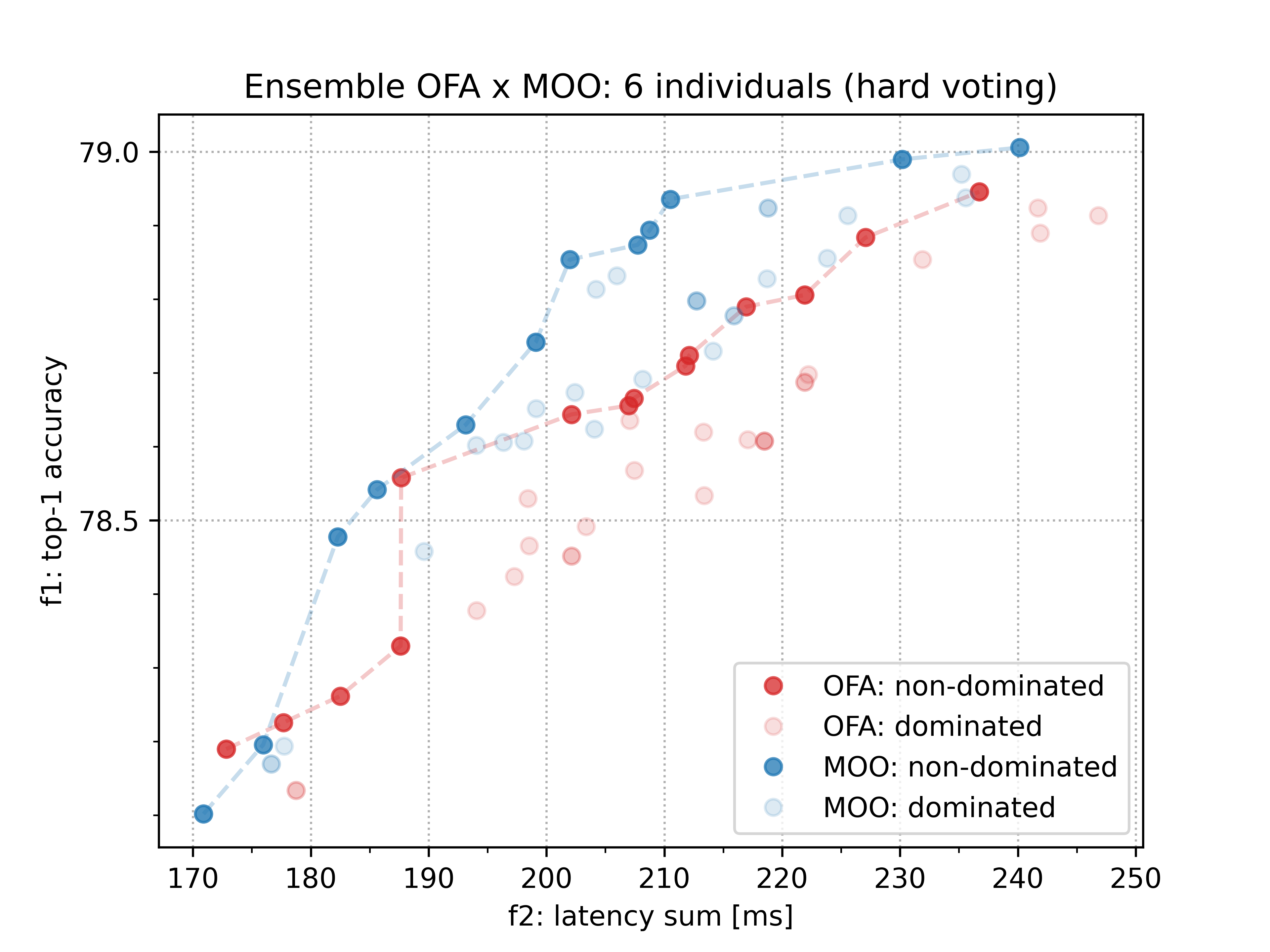}
\caption{\label{fig:ensemble_moo_ofa}Comparison between ensembles with efficient and OFA searched architectures (individuals = 6, latency = sum, voting = hard).}
\end{figure}

The experiments with ensembles formed by the architectures found by the OFA search follow the second method of the ensembles with efficient architectures. That is, we take 43 different combinations of architectures for each of the ensembles with 2, 3, \ldots{}, 8 models in its compositions out of the 9 models available. Figure \ref{fig:ensemble_moo_ofa} illustrates a comparison between the ensembles with size 6, using the hard voting scheme and taking the sum of latencies of all models in the committee. Again we see that the ensemble with efficient architectures outperforms the ones with the architectures from the OFA search.

\section{Conclusion}
\label{sec:org5da2f9c}
We introduced the OFA², a technique that extends the OFA framework by formulating the search stage as a multi-objective optimization problem. Solving this problem in the multi-objective perspective can be seen as the same of decoupling the previously jointed search and deploy stages. While in the original OFA framework the user must make a decision a priori to the search with respect to the desired hardware constraint, now we alleviate this decision to be made a posteriori. Instead of searching for a unique architecture, now we get a pool of non-dominated solutions all at once at the end of the search procedure, each of them with a unique trade-off regarding the conflicting objective functions (accuracy and latency). Besides of improving the efficiency of the search stage, our method is also capable of finding better architectures with respect to the accuracy evaluated on the ImageNet dataset.
Furthermore, we conduct a series of experiments related to the selection of architectures to form ensembles. We show that employing these efficient and diverse candidate solutions as components for ensembles may improve even more the overall performance.
\section*{References}
\label{sec:org4b59872}
\begin{hangparas}{.25in}{1}
\hypertarget{citeproc_bib_item_1}{[1] F. Hutter, L. Kotthoff, and J. Vanschoren, Eds., \textit{Automated Machine Learning: Methods, Systems, Challenges}. Cham: Springer International Publishing, 2019. doi: \href{https://doi.org/10.1007/978-3-030-05318-5}{10.1007/978-3-030-05318-5}.}

\hypertarget{citeproc_bib_item_2}{[2] B. Zoph and Q. Le, “Neural Architecture Search with Reinforcement Learning,” May 2017. Accessed: Nov. 21, 2022. [Online]. Available: \url{https://openreview.net/forum?id=r1Ue8Hcxg}}

\hypertarget{citeproc_bib_item_3}{[3] B. Zoph, V. Vasudevan, J. Shlens, and Q. V. Le, “Learning Transferable Architectures for Scalable Image Recognition,” in \textit{2018 IEEE/CVF Conference on Computer Vision and Pattern Recognition}, Jun. 2018, pp. 8697–8710. doi: \href{https://doi.org/10.1109/CVPR.2018.00907}{10.1109/CVPR.2018.00907}.}

\hypertarget{citeproc_bib_item_4}{[4] E. Real \textit{et al.}, “Large-Scale Evolution of Image Classifiers,” in \textit{Proceedings of the 34th International Conference on Machine Learning}, Jul. 2017, pp. 2902–2911. Accessed: Nov. 10, 2022. [Online]. Available: \url{https://proceedings.mlr.press/v70/real17a.html}}

\hypertarget{citeproc_bib_item_5}{[5] E. Real, A. Aggarwal, Y. Huang, and Q. V. Le, “Regularized Evolution for Image Classifier Architecture Search,” \textit{Proceedings of the aaai conference on artificial intelligence}, vol. 33, no. 01, pp. 4780–4789, Jul. 2019, doi: \href{https://doi.org/10.1609/aaai.v33i01.33014780}{10.1609/aaai.v33i01.33014780}.}

\hypertarget{citeproc_bib_item_6}{[6] H. Liu, K. Simonyan, and Y. Yang, “DARTS: Differentiable Architecture Search,” 2018, doi: \href{https://doi.org/10.48550/ARXIV.1806.09055}{10.48550/ARXIV.1806.09055}.}

\hypertarget{citeproc_bib_item_7}{[7] A. Vaswani \textit{et al.}, “Attention is All you Need,” in \textit{Advances in Neural Information Processing Systems}, 2017, vol. 30. Accessed: Jan. 17, 2023. [Online]. Available: \url{https://papers.nips.cc/paper/2017/hash/3f5ee243547dee91fbd053c1c4a845aa-Abstract.html}}

\hypertarget{citeproc_bib_item_8}{[8] D. So, Q. Le, and C. Liang, “The Evolved Transformer,” in \textit{Proceedings of the 36th International Conference on Machine Learning}, May 2019, pp. 5877–5886. Accessed: Jan. 17, 2023. [Online]. Available: \url{https://proceedings.mlr.press/v97/so19a.html}}

\hypertarget{citeproc_bib_item_9}{[9] E. Strubell, A. Ganesh, and A. McCallum, “Energy and Policy Considerations for Modern Deep Learning Research,” \textit{Proceedings of the aaai conference on artificial intelligence}, vol. 34, no. 09, pp. 13693–13696, Apr. 2020, doi: \href{https://doi.org/10.1609/aaai.v34i09.7123}{10.1609/aaai.v34i09.7123}.}

\hypertarget{citeproc_bib_item_10}{[10] H. Cai, C. Gan, T. Wang, Z. Zhang, and S. Han, “Once for All: Train One Network and Specialize it for Efficient Deployment,” Apr. 2020. Accessed: Nov. 08, 2022. [Online]. Available: \url{https://iclr.cc/virtual_2020/poster_HylxE1HKwS.html}}

\hypertarget{citeproc_bib_item_11}{[11] J. Deng, W. Dong, R. Socher, L.-J. Li, K. Li, and L. Fei-Fei, “ImageNet: A large-scale hierarchical image database,” in \textit{2009 IEEE Conference on Computer Vision and Pattern Recognition}, Jun. 2009, pp. 248–255. doi: \href{https://doi.org/10.1109/CVPR.2009.5206848}{10.1109/CVPR.2009.5206848}.}

\hypertarget{citeproc_bib_item_12}{[12] H. Cai, L. Zhu, and S. Han, “ProxylessNAS: Direct Neural Architecture Search on Target Task and Hardware,” Feb. 2022. Accessed: Dec. 12, 2022. [Online]. Available: \url{https://openreview.net/forum?id=HylVB3AqYm}}

\hypertarget{citeproc_bib_item_13}{[13] E. K. Burke and G. Kendall, Eds., \textit{Search Methodologies}. Boston, MA: Springer US, 2014. doi: \href{https://doi.org/10.1007/978-1-4614-6940-7}{10.1007/978-1-4614-6940-7}.}

\hypertarget{citeproc_bib_item_14}{[14] K. Deb, A. Pratap, S. Agarwal, and T. Meyarivan, “A fast and elitist multiobjective genetic algorithm: NSGA-II,” \textit{Ieee transactions on evolutionary computation}, vol. 6, no. 2, pp. 182–197, Apr. 2002, doi: \href{https://doi.org/10.1109/4235.996017}{10.1109/4235.996017}.}

\hypertarget{citeproc_bib_item_15}{[15] “SMS-EMOA: Multiobjective selection based on dominated hypervolume - ScienceDirect.” \url{https://www.sciencedirect.com/science/article/pii/S0377221706005443?via\%3Dihub} (accessed Jan. 19, 2023).}

\hypertarget{citeproc_bib_item_16}{[16] J. Blank and K. Deb, “Pymoo: Multi-Objective Optimization in Python,” \textit{Ieee access}, vol. 8, pp. 89497–89509, 2020, doi: \href{https://doi.org/10.1109/ACCESS.2020.2990567}{10.1109/ACCESS.2020.2990567}.}

\hypertarget{citeproc_bib_item_17}{[17] C. R. Harris \textit{et al.}, “Array programming with NumPy,” \textit{Nature}, vol. 585, no. 7825, 7825, pp. 357–362, Sep. 2020, doi: \href{https://doi.org/10.1038/s41586-020-2649-2}{10.1038/s41586-020-2649-2}.}

\hypertarget{citeproc_bib_item_18}{[18] A. Paszke \textit{et al.}, “PyTorch: An Imperative Style, High-Performance Deep Learning Library,” in \textit{Advances in Neural Information Processing Systems}, 2019, vol. 32. Accessed: Jan. 19, 2023. [Online]. Available: \url{https://papers.nips.cc/paper/2019/hash/bdbca288fee7f92f2bfa9f7012727740-Abstract.html}}
\end{hangparas}
\end{document}

%% file: def.tex
\def \BibTeX{{\rm B\kern-.05em{\sc i\kern-.025em b}\kern-.08em T\kern-.1667em\lower.7ex\hbox{E}\kern-.125emX}}

%% file: author.tex
\author{
  \IEEEauthorblockN{Rafael C. Ito}
  \IEEEauthorblockA{
    \textit{School of Electrical and Computer Engineering} \\
    \textit{University of Campinas (Unicamp)}\\
    Campinas, Brazil \\
    ito.rafael@gmail.com
  }

  \and

  \IEEEauthorblockN{Fernando J. Von Zuben}
  \IEEEauthorblockA{
    \textit{School of Electrical and Computer Engineering} \\
    \textit{University of Campinas (Unicamp)}\\
    Campinas, Brazil \\
    vonzuben@unicamp.br
  }
}